\documentclass[10pt,twocolumn,letterpaper]{article}
\usepackage[pagenumbers]{mystyle}
\usepackage{bm}
\usepackage{multirow}
\usepackage{array}
\usepackage{booktabs}
\usepackage{makecell}
\usepackage{tcolorbox}

\usepackage{amsfonts}
\usepackage{nicefrac}
\usepackage{microtype}
\usepackage{xcolor}
\usepackage{amsmath}
\usepackage[pagebackref,breaklinks,colorlinks,citecolor=green]{hyperref}

\newcommand{\authorskip}{\hspace{5mm}}

\title{Towards Improving Document Understanding: An Exploration \protect\\ on Text-Grounding via MLLMs}

\author{
	Yonghui Wang$^{1}$
	\authorskip Wengang Zhou$^{1*}$
	\authorskip Hao Feng$^{1}$
	\authorskip Keyi Zhou$^{1}$
	\authorskip Houqiang Li$^{1}$\thanks{Corresponding authors: Wengang Zhou and Houqiang Li.} \\ [1mm]
	{
		\fontsize{10.4pt}{9.84pt}\selectfont
		\textsuperscript{1} University of Science and Technology of China \hspace{5.7mm}
	} \\ [0mm]
	{
		\fontsize{9.4pt}{9.84pt}\selectfont
            \{wyh1998, haof, kyzhou2000\}@mail.ustc.edu.cn, \{zhwg, lihq\}@ustc.edu.cn
        }
}

\begin{document}
\maketitle

\begin{abstract}
In the field of document understanding, significant advances have been made in the fine-tuning of Multimodal Large Language Models (MLLMs) with instruction-following data.
Nevertheless, the potential of text-grounding capability within text-rich scenarios remains underexplored.
In this paper, we present a text-grounding document understanding model, termed TGDoc, which addresses this deficiency by enhancing MLLMs with the ability to discern the spatial positioning of text within images.
Empirical evidence suggests that text-grounding improves the model's interpretation of textual content, thereby elevating its proficiency in comprehending text-rich images.
Specifically, we compile a dataset containing 99K PowerPoint presentations sourced from the internet.
We formulate instruction tuning tasks including text detection, recognition, and spotting to facilitate the cohesive alignment between the visual encoder and large language model. 
Moreover, we curate a collection of text-rich images and prompt the text-only GPT-4 to generate 12K high-quality conversations, featuring textual locations within text-rich scenarios.
By integrating text location data into the instructions, TGDoc is adept at discerning text locations during the visual question process.
Extensive experiments demonstrate that our method achieves state-of-the-art performance across multiple text-rich benchmarks, validating the effectiveness of our method.
The source code is publicly available at \url{https://github.com/harrytea/TGDoc}.
\end{abstract}

\section{Introduction}
\label{sec:intro}
With the advancement for fine-tuning with instruction-following data, large language models have showcased exceptional generalization capabilities across various downstream tasks. 
Recently, models such as Flamingo~\cite{alayrac2022flamingo}, BLIP-2~\cite{li2023blip}, LLaVA~\cite{liu2023visual}, and MiniGPT-4~\cite{zhu2023minigpt} have extended these capabilities into multimodal domains, achieving the integrated visual-language understanding.
However, due to the absence of instruction-following data pertinent to text-rich scenarios, these models are unable to identify and comprehend the text within images, thereby exhibiting significant limitations in the processing of text-rich images.

\begin{figure}[t]
	\centering
	\includegraphics[width=0.98\linewidth]{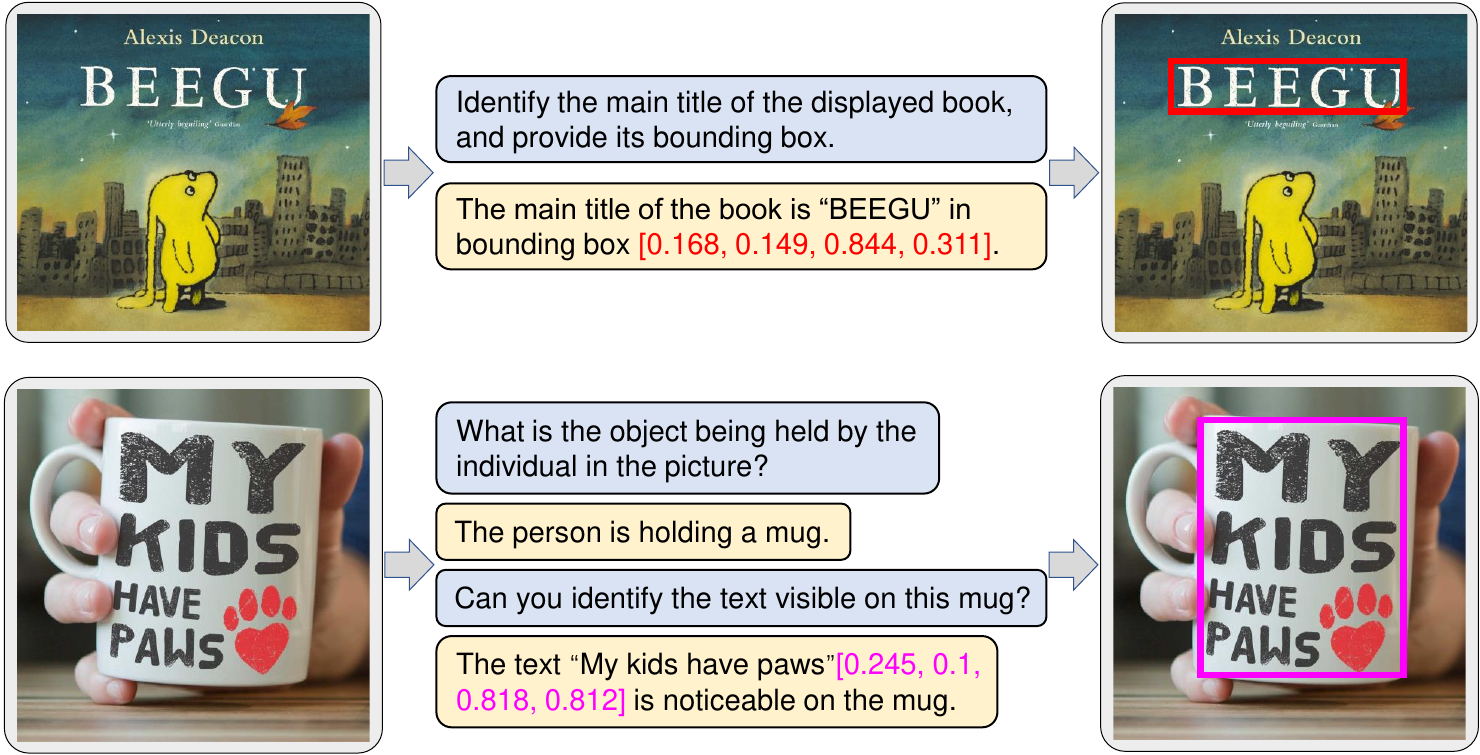}
	\caption{VQA in text-rich images with text-grounding. Given an image and a query, the model generates responses in an autoregressive way, and concurrently outputs the basis of reasoning, \emph{i.e.}, the bounding box that delineates the location of the answer.}
	\label{fig:intro}
\end{figure}

To address this issue, LLaVAR~\cite{zhang2023llavar} prompts Optical Character Recognition (OCR) results and image caption to text-only GPT-4, thereby creating a rich instruction-following dataset enriched with textual content. 
Following this, mPLUG-DocOwl~\cite{ye2023mplugdoc} harvests diverse forms of document data to augment the model's perception to textual features. 
Concurrently, UniDoc~\cite{feng2023unidoc} performs instruction tuning with OCR-based tasks, equipping the model with comprehensive abilities in text detection, recognition, and spotting, thereby deepening document understanding.
Although these methods have achieved impressive performance, the potential of text grounding as an instrument for document understanding is still less explored.

UniDoc~\cite{feng2023unidoc} has demonstrated that the use of spotting instruction templates can enhance the performance of detection and recognition tasks by incorporating the explicit location information.
Furthermore, empirical evidence from Shikra~\cite{chen2023shikra} indicates that the inclusion of center point locations for objects can mitigate hallucinations and improve the precision of its responses.
Similarly, we argue that text-grounding plays an important role in text-rich VQA tasks.
As illustrated in~\cref{fig:intro}, we aim to develop a model that can accurately respond to a question associated with an image and simultaneously identify and clarify the specific text region within the image that corresponds to the rationale behind the answer.
Generally, text-grounding ability has several benefits: 
First, it provides a basis for reasoning.
By incorporating text-grounding into VQA tasks, models accurately focus on image regions that are related to the question, thus leading to better performance. 
Second, it strengthens the interpretability of models, making the generated answers more convincing. 
Third, in text-rich scenarios where textual information are interleaved and complicated, the introduction of text-grounding elevates the interactive experience for users. 
We contend that text-grounding can further excavate the model's potential and release the extensive world knowledge contained within the model.

To endow the TGDoc model with text-grounding capability, we accumulate a large amount of instruction-following data enriched with text location details.
During the pre-training stage, we curate 99K PowerPoint presentations from the internet for image-text alignment, cultivating the model's proficiency in text detection and recognition across diverse scenarios.
In the fine-tuning stage, we prompt text-only GPT-4 with OCR results and image captions to produce 12K high-quality, multi-turn conversations. 
Additionally, we incorporate all data used by LLaVAR~\cite{zhang2023llavar} during both two stages. 
Experimental results indicate that using instruction tuning with the collected instruction-following data, our method exhibits improved performance on understanding text-rich scenes and has text-grounding capability. 

Our contributions can be summarized as follows:

\begin{itemize}
	\item We conduct an in-depth exploration of text-grounding in MLLMs without introducing extra detection modules, and substantiate its critical role in processing and understanding text-rich images.
	To the best of our knowledge, this is the first investigation into enhancing document understanding by using text-grounding technique.
	\item We develop an instructing-following dataset comprising 99K PowerPoint slides and 12K high-quality conversations, with each annotated with location of text.
	\item Extensive experiments show that our method achieves the state-of-the-art results on several text-rich VQA benchmarks, validating the effectiveness of our method.
\end{itemize}

\section{Related Work}
\label{sec:relate}

In this section, we begin with a brief overview of recent developments in multimodal large language models.
We then introduce their applications in document understanding and conclude by discussing the efforts to integrate multimodal large language models with grounding capabilities.

\subsection{Multimodal Large Language Models}
Large language models through instruction tuning have demonstrated impressive zero-shot capabilities for new tasks~\cite{ouyang2022training,wang2022self,chung2022scaling,xu2023baize,touvron2023llama}.
In particular, LLaMA~\cite{touvron2023llama}, a prominent open-source model, has drawn attention in research initiatives like Alpaca~\cite{alpaca2023} and Vicuna~\cite{chiang2023vicuna}, which utilize the generated instruction-following data to further excavate the potential of the model.
However, it is noteworthy that these studies exclusively accept the text as input.

Recently, researchers have extended the instruction tuning to the multimodal domain, particularly with a focus on images.
Specifically, Flamingo~\cite{alayrac2022flamingo} and BLIP-2~\cite{li2023blip} have pioneered visual and language integration by constructing diverse image-text alignment modules, laying a foundation for future research.
Similar to BLIP-2~\cite{li2023blip}, MiniGPT-4~\cite{zhu2023minigpt} utilizes a visual encoder that integrates ViT-G/14~\cite{fang2023eva} with Q-former and a linear projection layer to link the visual features with the large language model.
To mitigate pre-training challenges like word repetition and irrelevant content, the authors leverage ChatGPT for data refinement.
LLaVA~\cite{liu2023visual} employs a simple linear layer as a bridge to project image features into word embedding space.
Concurrently, mPLUG-Owl~\cite{ye2023mplug} utilizes a visual abstractor coupled with cross-attention to align visual representations with the large language model.
Recently, GPT-4V~\cite{yang2023dawn} has shown unprecedented proficiency in handling intricate multi-modal inputs. 
As a robust multi-modal system, GPT-4V possesses a profound ability to interpret visual information, significantly elevating human-machine interactions.

However, beyond GPT-4V~\cite{yang2023dawn}, the above efforts face difficulties in processing images within text-rich contexts. 
This challenge primarily arises from an insufficiency of relevant data, such as OCR-based or other text-related instruction-following data.
To address this issue, many related works have subsequently emerged~\cite{zhang2023llavar,ye2023mplugdoc,feng2023unidoc,ye2023ureader}.

\subsection{Document Understanding with MLLMs}
Recently, with the impressive zero-shot performance of multimodal large language models in visual-language tasks~\cite{tiong2022plug,driess2023palm,pmlrv162li22n}, the field of document understanding has shifted from supervised methods~\cite{xu2020layoutlm,wang2021layoutreader,xu2020layoutlmv2,yu2023structextv2} to generative approaches~\cite{zhang2023llavar,ye2023mplugdoc}, achieving several notable results.

LLaVAR~\cite{zhang2023llavar} leverages GPT-4~\cite{openai2023gpt4} with OCR tools to produce abundant instruction-following data for text-rich images.
Specifically, during the pre-training phase, they utilize the noisy instruction-following data to align image and text, enabling the model with OCR capability.
During fine-tuning, they employ a text-only GPT-4 to generate high-quality multi-turn conversations. 
Through instruction tuning, the model exhibits a marked performance on diverse text-based VQA datasets~\cite{mishra2019ocr,biten2019icdar,singh2019towards,mathew2021docvqa}, surpassing LLaVA~\cite{liu2023visual}.
mPLUG-DocOwl~\cite{ye2023mplugdoc}, an evolution of mPLUG-Owl~\cite{ye2023mplug}, is optimized for text-rich scenarios.
By creating diverse document instructional data, it excels in several downstream tasks within an OCR-free environment.
UniDoc~\cite{feng2023unidoc} mitigates potential data distribution inconsistencies between pre-training and fine-tuning by employing PowerPoint presentations to generate a large corpus of OCR-related instruction-following data.
By doing so, UniDoc synthesizes multi-tasks into a unified framework, achieving leading results on several benchmarks. 
Moreover, notable studies like UReader~\cite{ye2023ureader} and KOSMOS-2.5~\cite{lv2023kosmos} have undertaken diverse explorations in document understanding.

\subsection{Grounding Ability in MLLMs}
The exploration of grounding capabilities within MLLMs has increasingly attracted many researchers' interest.
Shikra~\cite{chen2023shikra} introduces a new referential dialogue task and reorganizes existing datasets with bounding boxes to incorporate object positional details into the instruction-following data.
After fine-tuning, the model exhibits impressive grounding ability without needing extra detection modules.
Moreover, the authors observe that the positional information in VQA tasks effectively reduces visual hallucinations.
KOSMOS-2~\cite{peng2023kosmos} has assembled a set of grounded image-text pairs.
After training, the model can perceive objects within images.
Ferret~\cite{you2023ferret} offers a unique method for object grounding in scenes, capable of recognizing any shape and granular target objects in images by using an innovative hybrid regional representation strategy.
The trained model is proficient in recognizing points, bounding boxes, and free-form shapes regional representations.

In text-rich scenarios, the grounding capabilities are still less explored. 
In this study, we focus on exploring the grounding of text within multimodal large language models. 
Our empirical results indicate that incorporating text-grounding ability can boost the understanding and interpretability of the model in text-rich scenarios.

\section{Method}
In this section, we initially outline the architecture of our TGDoc model and then provide a detailed description of how text-grounding is represented in our dataset.
Subsequently, we describe a comprehensive process involved in creating our grounded instruction-following dataset, encompassing both the pre-training and fine-tuning stages.

\subsection{Model Architecture}

\begin{figure}[t]
	\centering
	\includegraphics[width=0.98\linewidth]{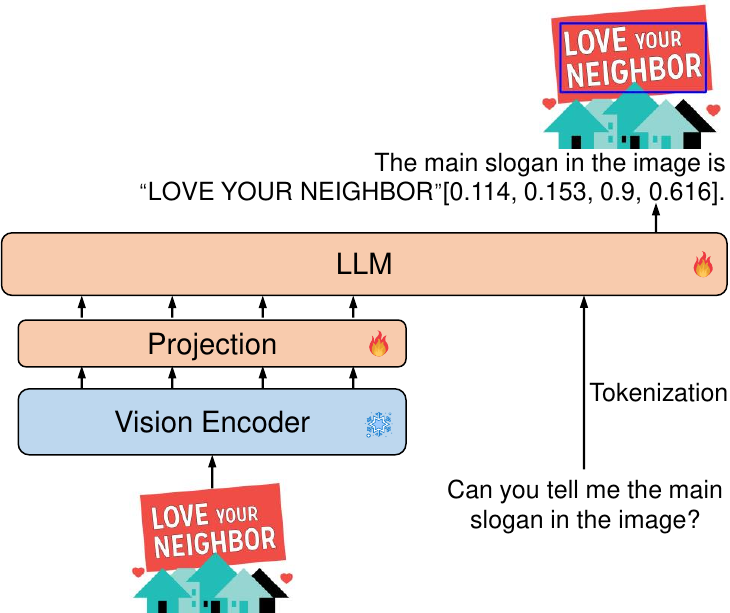}
	\caption{Overall architecture of TGDoc. It consists of a vision encoder, initialized with CLIP~\cite{radford2021learning} weights, a linear projection layer, and the Vicuna~\cite{chiang2023vicuna} large language model. Given an image and a query, the model will generate the answer while providing its reasoning textual regions.}
        \vspace{-0.2cm}
	\label{fig:method}
\end{figure}


\begin{table*}[t!]\centering
	\begin{minipage}{2.0\columnwidth}\vspace{0mm} \centering
		\begin{tcolorbox} 
			\centering
			\footnotesize
			\begin{tabular}{p{0.97\columnwidth} c}
				{\bf Captions generated by BLIP-2}  & \hspace{-4.5cm} \multirow{5}{*}{ \includegraphics[height=4.5cm]{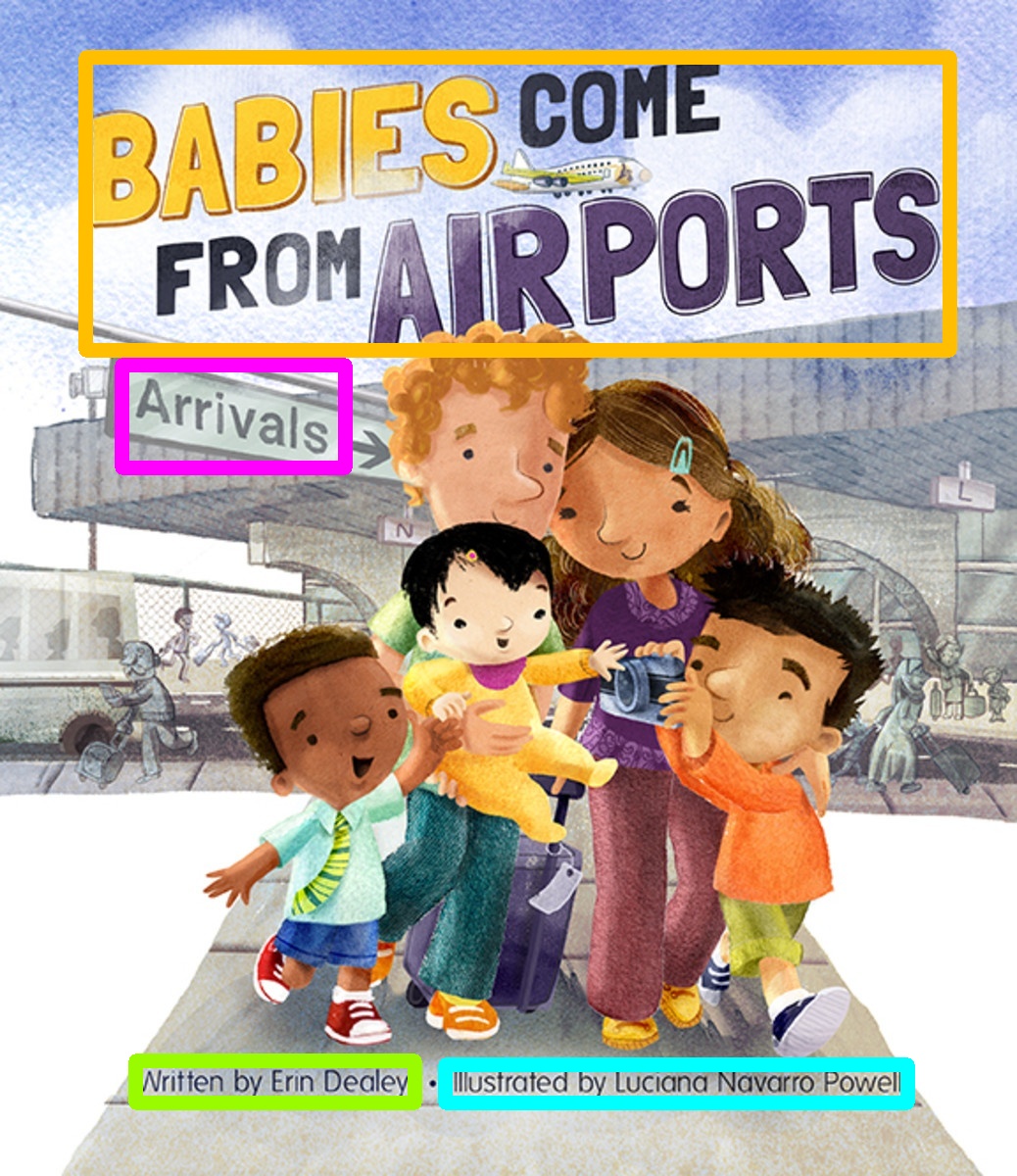}} \\
				``babies are from airports'', & \\
				``babies come from airports'', & \\
				``the cover of babies come from airports''. & \\
				{\bf PaddleOCR Results}  & \\
				``BABIES''[0.084,0.067,0.496,0.190], 
				``COME''[0.504,0.049,0.711,0.130], & \\
				``FROM''[0.150,0.193,0.367,0.270], ``AIRPORTS''[0.350,0.138,0.934,0.298], & \\
				``Arrivals=''[0.128,0.318,0.381,0.400], & \\
				``Written by Erin Dealey lllustrated by Luciana Navarro Powel''[0.136,0.909,0.887,0.930]. & \\
				{\bf EasyOCR Results}  & \\
				``BABI-S CONE''[0.063,0.032,0.733,0.208], 
				``FROM''[0.144,0.186,0.371,0.279], & \\ 
				``AIRPORTS''[0.353,0.124,0.940,0.304],
				``Arrivals''[0.120,0.310,0.340,0.398], & \\ 
				``Writien''[0.133,0.907,0.225,0.933], 
				``by ''[0.219,0.902,0.268,0.939], & \\
				``Erin Dealey''[0.261,0.902,0.408,0.938], & \\
				``Illustrated by Luciana Navarro Powell''[0.438,0.905,0.893,0.938].  & \\
				\hrulefill & \\
				{\bf Final Conversation}  & \\
				Q1: Identify the main title displayed in the image. & \\
				A1: The main title in the image is “BABIES COME FROM AIRPORTS”[0.084, 0.049, 0.934, 0.298]. & \\
				Q2: Are there any distinct sections denoted in the background? If so, please specify and support your response with the text and its bounding box. & \\
				A2: Yes, it seems that the babies are from airports and the airport stop says “Arrivals”[0.12, 0.31, 0.34, 0.398]. & \\
				Q3: Can you name the creators of the content depicted in the image? Provide the relevant text and its bounding box for justification. & \\
				A3: The creators are “Written by Erin Dealey”[0.133, 0.902, 0.408, 0.938] and “Illustrated by Luciana Navarro Powell”[0.438, 0.905, 0.893, 0.938]. & \\
			\end{tabular}
		\end{tcolorbox}
		\vspace{-2mm}
		\caption{The construction of the instruction-following dataset during the fine-tuning stage. Given an image, we employ BLIP-2~\cite{li2023blip} to produce three captions. In parallel, we extract text and bounding boxes from the image using two OCR engines. These captions and OCR results are then combined to form prompts, guiding GPT-4 in generating multi-turn conversations. To maintain data quality, these conversations are manually reviewed and refined, removing any content unrelated to the image.}
		\label{tab:gpt4_generate}
	\end{minipage}
\end{table*}

As shown in~\cref{fig:method}, following LLaVA~\cite{liu2023visual}, the overall architecture of our TGDoc model consists of three components: a vision encoder, a linear projection layer, and a large language model.
Specifically, we employ the pre-trained CLIP-ViT-L/14~\cite{radford2021learning} as vision encoder, configured to process images at two resolutions: $224\times 224$ and $336\times 336$.
The alignment module contains a single projection layer and is designed to transform the image features into the word embedding space compatible with the language decoder.
For the large language model, we select the Vicuna-7B~\cite{chiang2023vicuna}, an instruction-tuned variant based on LLaMA~\cite{touvron2023llama} to enhance our language understanding capabilities.

More specifically, given an image $\bm{I}^{H\times W\times C}$, we initially process it using a vision encoder to extract the image features $\bm{F}^{256\times 1024}$.
In this paper, we opt for grid features before the last transformer layer to serve as our image features.
These features, as opposed to those from the last layer, exhibit an enhanced ability to discern intricate details in the image~\cite{liu2023visual}.
This attribute is especially beneficial for the recognition of text in text-rich environments.
Subsequently, these image features $\bm{F}^{256\times 1024}$ are transformed through a linear projection layer into the word embedding space, resulting in transformed features $\bm{F}_{trans}^{256\times 4096}$ to align with the language decoder.
Moreover, user queries are tokenized within the same word embedding space and seamlessly concatenated with the visual embeddings to form uniform input vectors for the large language model.
Consistent with the previous methods~\cite{zhang2023llavar,ye2023mplugdoc,feng2023unidoc,chen2023shikra}, we continue to train the model with the next-token prediction task, \emph{i.e.}, generating the next token based on the sequence of preceding tokens.


\subsection{Grounded Input Representations}
Following the expression of shikra~\cite{chen2023shikra}, we adopt the natural language formatted notation $[x_{min},y_{min},x_{max},y_{max}]$ to represent the bounding boxes coordinates.
In this context, $[x_{min}, y_{min}]$ signifies the top-left corner of the text's minimum bounding rectangle, while $[x_{max},y_{max}]$ corresponds to the bottom-right corner.
We normalize the bounding boxes to $[0, 1]$ relative to the image size and maintain a three-decimal precision for each coordinate.
The bounding box notion can be integrated seamlessly into both the prompts and responses, encapsulated as ``\textless text\textgreater$[x_{min},y_{min},x_{max},y_{max}]$''.
For instance, as illustrated in~\cref{fig:method}, the structured input and output are presented as follows: ``USER: \textless image\textgreater Image Embeddings\textless /image\textgreater Can you tell me the main slogan in the image? ASSISTANT: The main slogan in the image is ``LOVE YOUR NEIGHBOR''[0.114, 0.153, 0.9, 0.616].''.
We employ the model to predict the coordinates in a manner analogous to natural language prediction, without introducing additional positional tokens or detection modules.

\subsection{Instruction-following Data for Pre-training}
We indicate that the data collection criteria stipulate not only textual richness but also diversity in content.
Inspired by UniDoc~\cite{feng2023unidoc}, we select PowerPoint slides as our instruction-following training data, due to their multiple advantages. 
Firstly, the slides are rich in structured text, featuring diverse fonts and artistic texts.
Unlike text from natural scenes, the content in PowerPoint slides is clearer and more organized, which markedly improves its compatibility with Optical Character Recognition (OCR) technique.
This clarity reduces the need for manual corrections, thus simplifying the data preprocessing phase.
Secondly, PowerPoint files encompass various non-textual elements like graphs and flowcharts,  which enrich the dataset with a diversity of image types.
Thirdly, the slides often contain interleaved visual and text information, such as photographs of scenes and photos of products, embedding the text within a broader narrative context, which is conducive to the model's understanding of complex image-text scenarios.

\begin{figure}[t]
	\centering
	\includegraphics[width=0.98\linewidth]{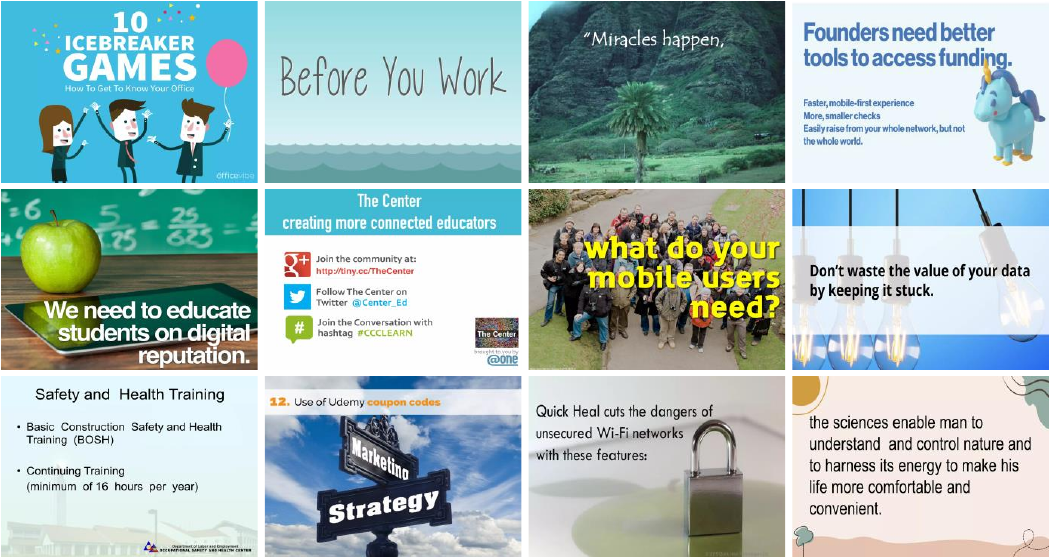}
	\caption{Examples of PowerPoint presentations collected by us.}
	\label{fig:ppt_data}
        \vspace{-0.2cm}
\end{figure}

\begin{figure}[t]
	\centering
	\includegraphics[width=0.98\linewidth]{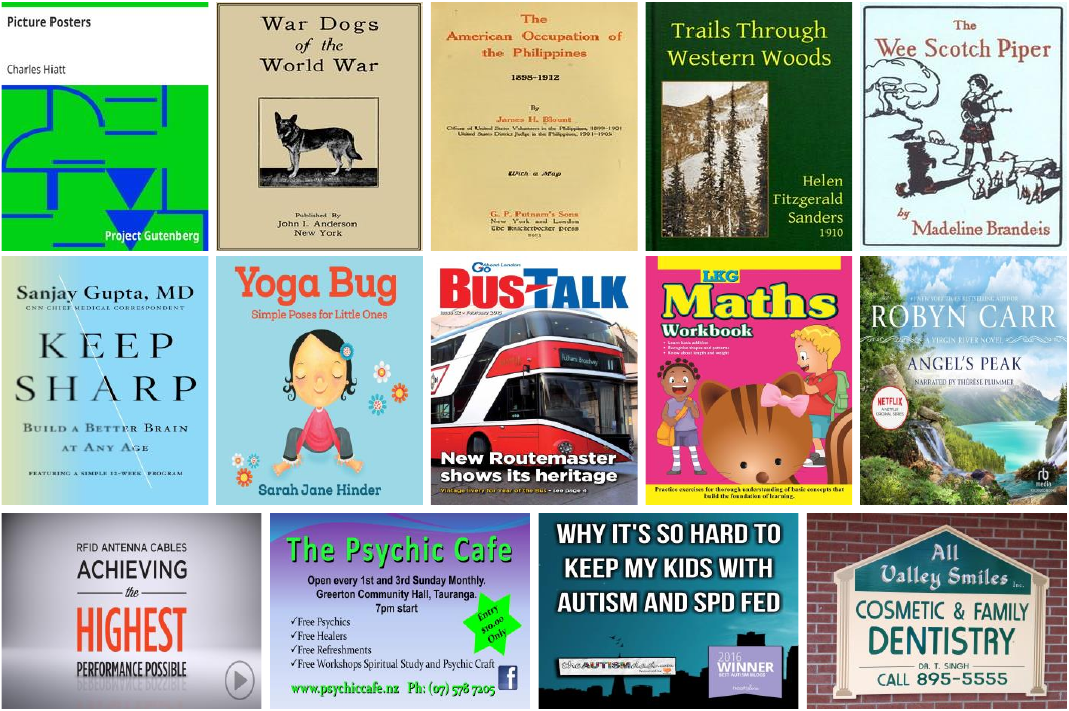}
	\caption{Examples of fine-tuning dataset collected by us. The first row consists of book covers curated from the internet, whereas the second and third rows display text-rich images sourced from the LAION-400M dataset~\cite{schuhmann2021laion}.}
	\label{fig:fine_data}
        \vspace{-0.2cm}
\end{figure}

In the construction of PowerPoint datasets, we consciously avoid automated generation techniques to prevent the risk of data homogeneity.
Instead, we chose to source data from the website, SlideShare~\footnote{https://www.slideshare.net\label{slideshare}}, a public platform for sharing presentations and multimedia materials.
Our collection spans various domains such as business, education, and athletics.
To refine the dataset, we apply the MD5 hashing algorithm~\cite{rivest1992md5} to eliminate any duplicates.
We then utilize PaddleOCR~\footnote{https://github.com/PaddlePaddle/PaddleOCR\label{paddle}}, an open-source tool, to extract both text and spatial bounding box data from each slide.
To ensure high data quality, we impose a threshold requiring that each slide must feature at least one block of text accounting for a minimum of 5\% of the slide's total image area.
This rigorous process yields data consisting of 99K image-text pairs, and we present several examples of slides in~\cref{fig:ppt_data}.

With respect to the accumulated PowerPoint slides data, we develop three distinct types of tasks for each image: recognition, detection, and spotting.
The instructions for these tasks are: ``Identifying all text along with its corresponding bounding box in the image'', ``Locating the box containing the word \textless text\textgreater'', and ``Please recognize and provide the text contained within the specified bounding box $[x_{min},y_{min},x_{max},y_{max}]$'', respectively. 
To broaden the scope of instructions, we utilize GPT-4 for query augmentation.
This approach enhances the diversity of instructional inputs, thereby bolstering the model's ability to generalize across a range of scenarios.

\subsection{Instruction-following Data for Fine-tuning}
To improve the model's proficiency in specific tasks and deepen its understanding and execution of standard Visual Question Answering (VQA) tasks, we carefully design a total of 12K multi-turn conversations.
Similarly to the pre-training stage, this dataset integrates textual content with associated bounding box.
The data comprises two components: book covers derived from public digital libraries and text-rich images from LAION-400M~\cite{schuhmann2021laion}.

\setlength{\tabcolsep}{10pt}
\begin{table}[t]
	\centering
	\resizebox{\linewidth}{!}{
		\begin{tabular}{l|cccc}
			\Xhline{2.5\arrayrulewidth}
			Data         & Instruction   & Conv & Que len & Ans len \\
			\hline
			LLaVA pre~\cite{liu2023visual}    & CC3M          & 595K  & 15.9  & 15.4    \\
			LLaVAR pre~\cite{zhang2023llavar}   & PaddleOCR     & 422K  & 17.2  & 48.8    \\
			Ours pre     & PaddleOCR+box & 99K   & 42.5  & 80.7    \\
			\hline
			LLaVA fine~\cite{liu2023visual}   & GPT-4         & 158K  & 15.9  & 32.4    \\
			LLaVAR fine~\cite{zhang2023llavar}  & GPT-4         & 16K   & 15.1  & 93.1    \\
			Ours fine    & GPT-4+box  & 12K   & 31.2  & 41.6    \\
			\Xhline{2.5\arrayrulewidth}
		\end{tabular}
	}
	\caption{Analysis of our data with LLaVA~\cite{liu2023visual} and LLaVAR~\cite{zhang2023llavar}. ``pre'' and ``fine'' are the data utilized in the pre-training and fine-tuning stage, respectively. ``Que len'' and ``Ans len'' represent the average number of tokens after the LLaMA~\cite{touvron2023llama} tokenization.}
	\label{tab:data_analysis}
\end{table}

\begin{table*}[t]
	\centering
	\setlength\tabcolsep{1pt}
	\small
	\resizebox{0.9\linewidth}{!}{%
		\begin{tabular}{lcccccccccc}
			\Xhline{2.5\arrayrulewidth}                                                  
			\multirow{3}{*}{Method} & \multicolumn{6}{c}{VQA}  & \multicolumn{3}{c}{KIE} & \multirow{3}{*}{\quad Avg.\quad}  \\ 
			\cmidrule(lr){2-7}  \cmidrule(lr){8-10} 
			& STVQA & OCRVQA & TextVQA & DocVQA  & InfoVQA & ChartQA & FUNSD & SROIE & POIE & \\
			\hline
			
			BLIP-2 OPT$\mathrm{_{6.7b}}$~\cite{li2023blip}  & 13.36 & 10.58 & 21.18 &  0.82 & 8.82 & 7.44 & 0.00   & 0.00  & 0.02  & 6.22  \\
			BLIP-2 FlanT5$\mathrm{_{XXL}}$~\cite{li2023blip}  & 21.70 & 30.74 & 32.18 &  4.86 & 10.17 & 7.20 & 1.19 & 0.20 & 2.52 &  12.31 \\
			OpenFlamingo~\cite{awadalla2023openflamingo} & 19.32 & 27.82 & 29.08 & 5.05 & \underline{14.99} & 9.12 & 0.85 & 0.12 & 2.12 &  12.05 \\
			LLaVA~\cite{liu2023visual} & 22.08 & 11.36 & 28.86 & 4.49 & 13.78 & 7.28 & 1.02 & 0.12 & 2.09 & 10.12 \\
			MiniGPT-4~\cite{zhu2023minigpt} & 14.02 & 11.52 & 18.72 & 2.97 & 13.32 &  4.32 & 1.19 & 0.04 & 1.31 & 7.49 \\
			mPLUG-Owl~\cite{ye2023mplug} & 29.26 & 28.62 & 40.28 & 6.88 & \textbf{16.46} & 9.52 & 1.02 & 0.64 & 3.26  & 15.10 \\  
			LLaVAR~\cite{zhang2023llavar} & 30.36 & 29.38 & 39.40 & 6.73  & 12.25 & 8.00 & 1.02 & 1.36 & 6.48   & 15.00 \\
			UniDoc~\cite{feng2023unidoc} & 30.78 & \underline{34.50} & 40.72 & 6.47 & 13.75 & 10.48 & 1.19 & 1.40 & 3.92 & 15.91 \\
			\hline
			TGDoc-224 & \underline{31.40} & 33.50 & \underline{41.86} & \underline{7.25} & 11.53 & \underline{11.74} & \textbf{1.70} & \underline{1.59} & \underline{9.08} & \underline{16.63} \\
			TGDoc-336 & \textbf{36.28} & \textbf{37.21} & \textbf{46.18} & \textbf{9.00} & 12.75 & \textbf{12.72} & \underline{1.36} & \textbf{3.00} & \textbf{22.16} & \textbf{20.07} \\
			\Xhline{2.5\arrayrulewidth}
	\end{tabular}}
	\caption{Quantitative comparison with previous multimodal large language models on Visual Question Answering (VQA) and Key Information Extraction (KIE) datasets. TGDoc-224 and TGDoc-336 represent the CLIP~\cite{radford2021learning} inputs at resolutions of $224\times 224$ and $336\times 336$. The best and the second results (accuracy \%) are highlighted in \textbf{bold} and \underline{underlined}, respectively.}
	\label{tab:quat}
\end{table*}

\noindent
\textbf{Book covers.}
We manually download 11K book covers from the openly accessible Project Gutenberg digital library~\footnote{https://gutenberg.org}.
For each cover, we carefully extract pivotal metadata, such as the title and author.
Subsequently, we utilize PaddleOCR~\textsuperscript{\ref{paddle}} to analyze each image, from which we reconstruct the text to attain the book's specifics along with their bounding boxes.
We utilize GPT-4 to design conversations revolving around the book's metadata, simultaneously requiring the model to explain its rationale behind the responses.
For example, a query could be ``Who is the author of this volume? Please justify your response with the bounding box $[x_{min},y_{min},x_{max},y_{max}]$.''
The model's response would include the author's name along with the bounding box coordinates, like: ``The work is authored by \textless text\textgreater$[x_{min},y_{min},x_{max},y_{max}]$.''

\noindent
\textbf{Text-rich scene images.}
We curate a corpus of 1K high-resolution (minimum $1024\times 1024$ pixels), text-rich images from LAION-400M dataset~\cite{schuhmann2021laion}.
Inspired by LLaVAR~\cite{zhang2023llavar}, we employ two OCR tools, PaddleOCR~\textsuperscript{\ref{paddle}} and EasyOCR\footnote{https://github.com/JaidedAI/EasyOCR} to process each image. 
In addition, we utilize BLIP-2~\cite{li2023blip} to generate three descriptions for every image.
Subsequently, as shown in~\cref{tab:gpt4_generate}, these elements, two OCR results along with the generated captions, are prompted to instruct GPT-4 in generating multi-turn conversations that concentrate on the textual content of image.
However, we observe some inconsistencies in the generated data, such as the unnecessary phrases like ``based on the paddleocr''. 
We manually review the conversation of each image to guarantee the quality, resulting in our final instruction fine-tuning dataset.

Finally, we exhibit the examples of our fine-tuning data in~\cref{fig:fine_data}.
Besides, we also provide an analysis of our data with that of LLaVA~\cite{liu2023visual} and LLaVAR~\cite{zhang2023llavar} in~\cref{tab:data_analysis}.

\section{Experiments}
\label{sec:exp}

\subsection{Implementation Details}
The implementation of TGDoc was executed on the Linux platform with eight A100 GPUs.
We incorporated the datasets from LLaVAR~\cite{zhang2023llavar} along with our own data.
For vision encoder, we selected CLIP-ViT-L/14~\cite{radford2021learning} and conducted experiments with image sizes of $224\times 224$ and $336\times 336$, respectively.
It has been observed that text-related tasks generally benefit from higher resolution~\cite{feng2023unidoc,ye2023mplugdoc,zhang2023llavar}.
We opted for Vicuna~\cite{chiang2023vicuna}, which is optimized for multi-tasks and an evolution of LLaMA~\cite{touvron2023llama} as our large language model.
For pre-training, only the linear projection layer was trained.
We applied the learning rate of 2e-3 and a batch size of 128.
While for fine-tuning, we adjusted both the linear projection layer and the large language model, reducing the learning rate to 2e-5 with a batch size of 32.
The AdamW optimizer~\cite{loshchilov2017decoupled} was utilized for parameter updates, with each stage undergoing one epoch of training.
Moreover, We employed a cosine annealing scheduler~\cite{loshchilov2016sgdr} and set the maximum sequence length limited to 2048. 

\subsection{Datasets and Evaluation Metrics}
To validate the effectiveness of text-grounding in enhancing document understanding, we evaluate our method on six text-based Visual Question Answering (VQA) datasets, including STVQA~\cite{biten2019icdar}, OCRVQA~\cite{mishra2019ocr}, TextVQA~\cite{singh2019towards}, DocVQA~\cite{mathew2021docvqa}, InfographicVQA~\cite{mathew2022infographicvqa}, and ChartQA~\cite{masry2022chartqa}. 
In addition, we also employ three key information extraction (KIE) datasets, including FUNSD~\cite{jaume2019funsd}, SROIE~\cite{huang2019icdar2019}, and POIE~\cite{kuang2023visual}.
For each question, we append the instruction ``Support your reasoning with the coordinates $[x_{min}, y_{min}, x_{max}, y_{max}]$'' at the end of the question to prompt the model to output the bounding box where the answer is located.
Following the evaluation protocol proposed by Liu \emph{et al.}~\cite{liu2023hidden}, we use accuracy metric to determine if the model-generated content correctly contains the ground truth answer.

\begin{figure*}[t]
	\centering
	\includegraphics[width=0.98\linewidth]{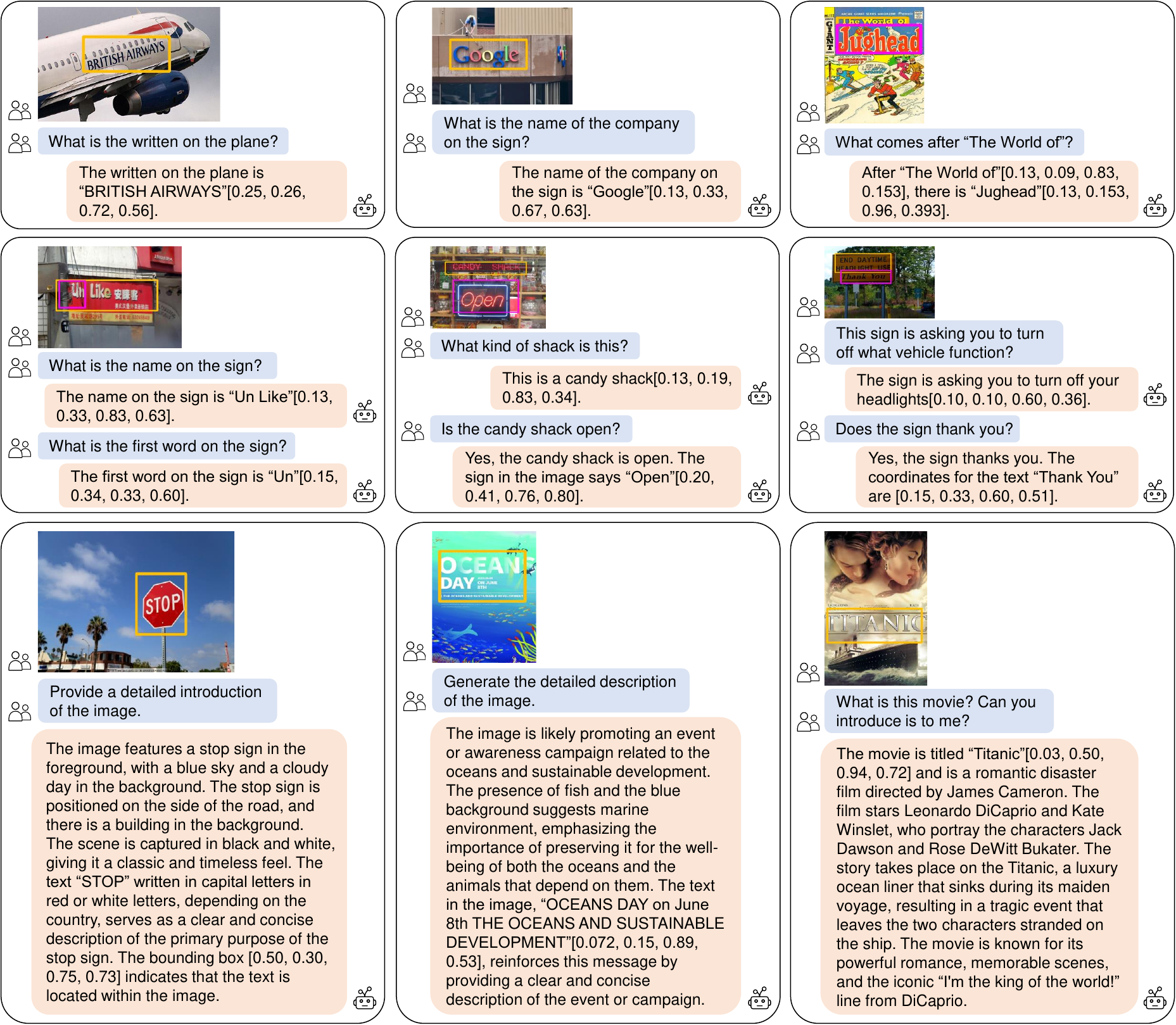}
	\caption{Visualization results of TGDoc. To conserve space, the bounding boxes generated by the model are visualized and integrated into the model's input. The results indicate that the model can accurately focus on regions pertinent to the answers, which notably enhances the model's interpretability.}
	\label{fig:qua1}
\end{figure*}

\subsection{Comparison with Other MLLMs}
\label{sec:5_3}

We compare our approach with recent multimodal large language models on Visual Question Answering (VQA) and Key Information Extraction (KIE) benchmarks. 
\cref{tab:quat} summarizes the quantitative comparison results.
We conduct experiments using the CLIP model~\cite{radford2021learning} at two resolutions: $224\times 224$ and $336\times 336$.
At the size of $224\times 224$, our method achieves state-of-the-art performance across several benchmarks, validating that introducing the text location information can boost the model's capacity for document understanding.
Increasing the image size to $336\times 336$ further enhances our results.
This conclusion aligns with previous methods~\cite{feng2023unidoc,ye2023mplugdoc,zhang2023llavar}, as higher resolution enriches the low-level details of textual content, aiding in image comprehension.
Although at an image resolution of $336\times 336$, our approach does not outperform other methods on the InfographicVQA benchmark~\cite{mathew2022infographicvqa}. 
We analyze that this benchmark comprises various complex infographics that require elementary reasoning and basic arithmetic skills, as answers cannot be readily extracted from the images.
Under these conditions, the utility of bounding boxes is restricted, and their requirement may potentially compound its reasoning workload.
We are considering a strategy that integrates chain-of-thought reasoning with bounding boxes to enhance performance in these challenging conditions for future research.

We also present the qualitative results of our approach in~\cref{fig:qua1}, complementing the previously reported quantitative results.
The model not only generates answers but also visually highlights the regions that substantiate its responses.
This implies that the model needs to identify the specific regions pertinent to the query, mirroring the human cognitive process of seeking relevant information to answer questions. 
However, it should be noted that our text-grounding is not always precise.
We speculate that this limitation may be attributed to the inherent characteristics of the frozen CLIP model~\cite{radford2021learning}, which prioritizes global semantics over meticulous attention to local textual details.

\subsection{Ablation Studies}
In this section, we conduct ablation studies on our curated pre-training and fine-tuning datasets, encompassing four distinct experiments.
The first experiment (labeled as ``Base'') excludes the data we collected, \emph{i.e.}, the same setting as LLaVAR~\cite{zhang2023llavar}. 
The second (labeled as ``Base+pre'') focuses on the impact of our pre-training data. 
The third (labeled as ``Base+fine'') investigates the influence of our fine-tuning data. 
To rigorously assess the impact of text-grounding on the model's comprehension ability, we undertake the fourth experiment (labeled as ``Ours+w/o box''), which removes all bounding boxes in data.
All experiments are uniformly performed with an input image size of $224\times 224$.

\begin{figure}[t]
	\centering
	\includegraphics[width=0.98\linewidth]{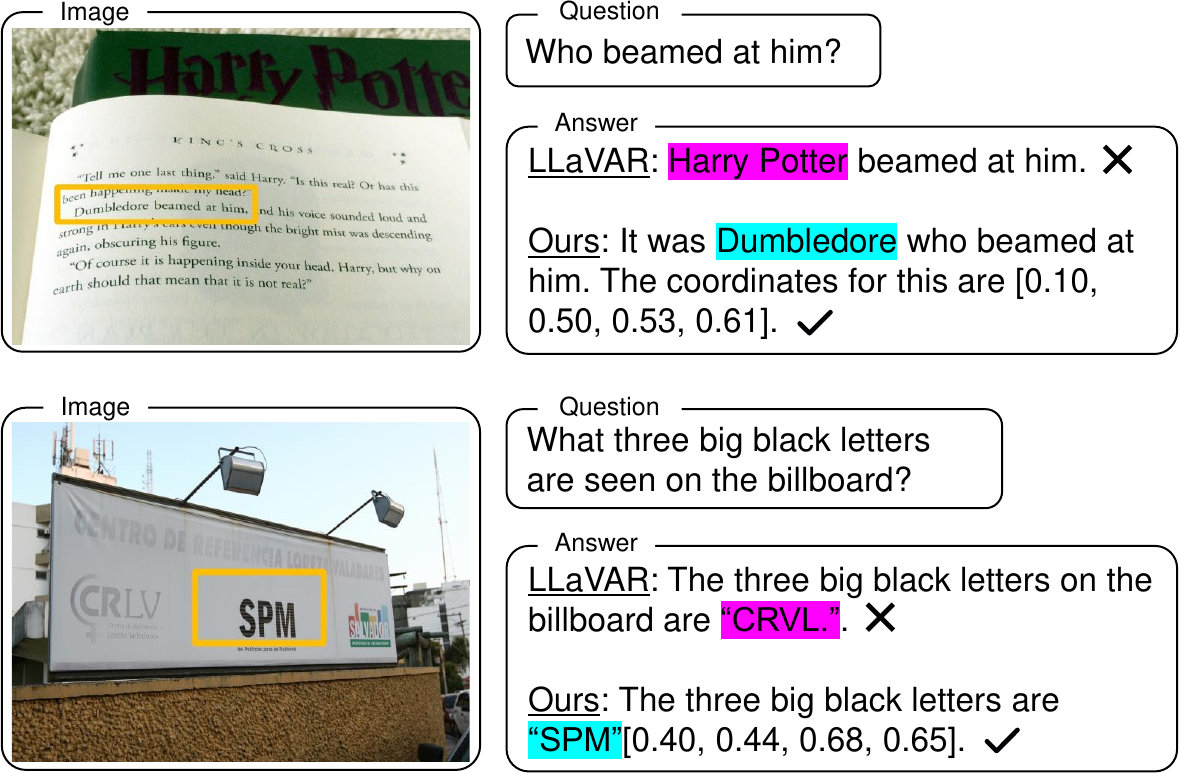}
	\caption{Examples of incorporating text-grounding capability. Compared to LLaVAR~\cite{zhang2023llavar}, our method delivers correct answers with enhanced accuracy.}
	\label{fig:abl_box_good}
\end{figure}

We present the results in~\cref{tab:abl_data}.
From the results, we can obtain two principal conclusions. 
First, although the volume of fine-tuning data is smaller than that of pre-training data, the performance based on fine-tuning data surpasses that based on pre-training.
We infer that incorporating specific grounded instruction tuning data during the fine-tuning stage enables the model to more accurately interpret user instructions. 
This leads to enhanced performance in various tasks such as Visual Question Answering (VQA). 
Additionally, this approach effectively facilitates the model's rapid acquisition of text-grounding skills, allowing it to focus on image regions directly pertinent to the answers.
Second, compared to the results without bounding boxes, the introduction of text-grounding further improves the performance.
As shown in~\cref{fig:abl_box_good}, compared to LLaVAR~\cite{zhang2023llavar}, our method delivers correct answers with enhanced accuracy.


\begin{table}[t]
	\centering
	\resizebox{\linewidth}{!}{
		\begin{tabular}{c|cc|cccc}
			\Xhline{2.5\arrayrulewidth}
			& \multicolumn{2}{c|}{Config} & \multicolumn{4}{c}{Benchmark}  \\
			\hline
			& $\mathcal{P}$ & $\mathcal{F}$ & textVQA & DocVQA & FUNSD  &  POIE  \\
			\hline
			Base         &        &                     & 39.40 & 6.73 & 1.02 & 6.48  \\
			Base+pre     & $\checkmark$ &               & 41.08 & 6.83 & 1.36 & 5.23  \\
			Base+fine    &  & $\checkmark$              & \underline{41.22} & 6.92 & \underline{1.53} & 7.04  \\    
			Ours+w/o box & $\checkmark$  & $\checkmark$ & 41.14 & \underline{7.06} & 1.02 & \underline{8.24}  \\
			Ours         & $\checkmark$  & $\checkmark$ & \textbf{41.86} & \textbf{7.25} & \textbf{1.70} & \textbf{9.08}  \\    
			\Xhline{2.5\arrayrulewidth}
		\end{tabular}
	}
	\caption{Ablation studies on our collected data. The best and the second results (accuracy \%) are highlighted in \textbf{bold} and \underline{underlined}, respectively. $\mathcal{P}$ denotes the data gathered for the pre-training stage, while $\mathcal{F}$ denotes the data for the fine-tuning stage.}
	\label{tab:abl_data}
\end{table}

\subsection{Discussion}
The experimental results substantiate that the text-grounding capability enables the model to concentrate more on regions within images relevant to the answers, thereby enhancing the model's comprehension in text-rich scenarios.
Furthermore, throughout the experiment, we also encounter a series of challenges, which will be discussed in the following.

\begin{figure}[t]
	\centering
	\includegraphics[width=0.98\linewidth]{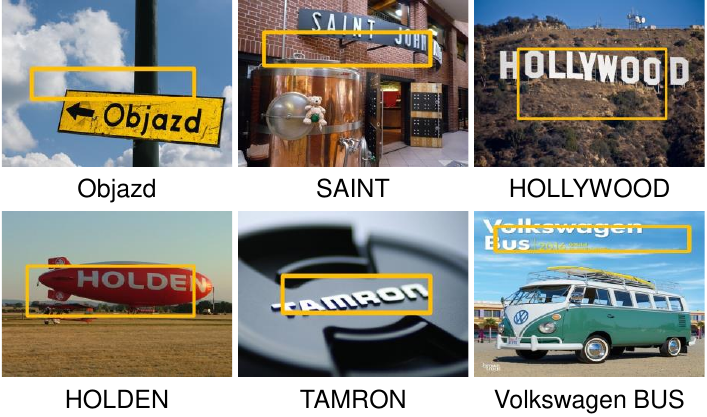}
	\caption{Some examples of imprecise bounding boxes produced by our method.}
	\label{fig:abl_box}
\end{figure}

\noindent
\textbf{Accuracy of bounding boxes.}
In~\cref{fig:abl_box}, we present some examples of inaccurate bounding boxes generated by our methods.
Compared to~\cref{fig:qua1}, these boxes either near the text or cover a fragment of it.
We speculate that this may stem from the capabilities of the visual encoder CLIP~\cite{radford2021learning}, which is based on ViT~\cite{dosovitskiy2020image} and trained primarily on images of natural objects, emphasizing the global features of images rather than optimizing for fine-grained local details like textual boundaries, which leads to challenges in text detection.
However, we argue that the bounding boxes, despite not always being exact, are vital for localization in VQA tasks.
They can further narrow down the model's retrieval scope, benefiting the overall understanding process.

\noindent
\textbf{Impact of resolution on the model.}
Image resolution significantly influences text-based tasks~\cite{zhang2023Monkey}.
Typically, text-rich images are of high resolution, and their compression to meet the CLIP's~\cite{radford2021learning} input sizes often leads to a loss of textual detail.
For the POIE dataset~\cite{kuang2023visual}, increasing the input resolution of CLIP~\cite{radford2021learning} from 224 to 336 pixels results in a 13\% accuracy improvement.
We point out that this is because most images in the POIE dataset~\cite{kuang2023visual} are low-resolution images, and a larger resolution preserves text details that might be lost at a lower dimension of 224.
Moreover, the integration of text-grounding has further amplified the model’s efficacy in processing the POIE dataset~\cite{kuang2023visual}.

\noindent
\textbf{Questions requiring multi-step reasoning.}
Our method is capable of providing positional information relevant to the answers within its responses. 
Nevertheless, as discussed in~\cref{sec:5_3}, for queries that require multi-step reasoning (where answers cannot be directly extracted from the image), the efficacy of text-grounding is constrained.
We speculate that this limitation arises due to the model focusing on irrelevant areas within incorrect bounding boxes, which not only leads to ineffective reasoning but also disrupts the entire reasoning process.
To address this issue, we consider adopting the chain-of-thought method~\cite{wei2022chain}, incorporating bounding boxes at each step of the inference process for future work.

\section{Conclusion}
Our research investigates text-grounding within document understanding, demonstrating its significant role in improving the model comprehension of text-rich scenes.
We create 99K PowerPoint slides for pre-training, focusing on detection, recognition, and spotting tasks for image-text alignment. 
For fine-tuning, we compile a dataset of 12K high-quality conversations with bounding box annotations to specify text locations. 
Experimental results with the collected grounded instruction-following dataset reveal that text-grounding enhances the model's interpretability and performance by identifying answer-related areas.
However, the model faces challenges in processing complex reasoning questions, which will be explored in our future research.

{
    \small
    \bibliographystyle{ieeenat_fullname}
    \bibliography{main}
}

\end{document}


\maketitle


\section*{A. Instruction Templates}
\label{sec:template}
\noindent
\textbf{Pre-training.}
For the collected 99K PowerPoint slides, we focus on three primary tasks: detection, recognition, and spotting.
For each task, we develop several instruction templates.
To enhance the model's capacity for interpreting diverse user instructions and to avoid overfitting to particular templates, we employ GPT-4 to further expand these templates. 
We present examples of the instructions in~\cref{tab:template}.

\begin{table*}[!ht]
	\centering
	\resizebox{1.0\linewidth}{!}{
		\begin{tabular}{c|l}
			\Xhline{2.5\arrayrulewidth}
			Task  &  Instruction template  \\
			\hline
			\multirow{10}{*}{detection} & Can you furnish the bounding box coordinates [xmin, ymin, xmax, ymax] for the text \textless text\textgreater in the image? \\
			& Please supply the coordinates [xmin, ymin, xmax, ymax] for the text \textless text\textgreater within the image. \\
			& Could you kindly provide the bounding box or coordinates [xmin, ymin, xmax, ymax] for the text \textless text\textgreater in the image? \\
			& I would like to know the coordinates [xmin, ymin, xmax, ymax] for the text \textless text\textgreater in the image, please. \\
			& Please identify the bounding box coordinates [xmin, ymin, xmax, ymax] for the text \textless text\textgreater within the image. \\
			& Can you please give me the coordinates [xmin, ymin, xmax, ymax] for the text \textless text\textgreater in the image? \\
			& Kindly provide the bounding box coordinates [xmin, ymin, xmax, ymax] for the text \textless text\textgreater within the image. \\
			& Please retrieve the coordinates [xmin, ymin, xmax, ymax] for the text \textless text\textgreater in the image. \\
			& I'm looking for the bounding box coordinates [xmin, ymin, xmax, ymax] for the text \textless text\textgreater in the image, could you provide them? \\
			& Please locate and share the bounding box coordinates [xmin, ymin, xmax, ymax] for the text \textless text\textgreater within the image. \\
			\hline
			\multirow{10}{*}{recognition} & Please recognize and supply the text enclosed within the given bounding box [xmin, ymin, xmax, ymax]. \\
			& Can you identify and provide the text that falls within the specified coordinates [xmin, ymin, xmax, ymax]? \\
			& I'd like you to detect and furnish the text contained within the provided bounding box [xmin, ymin, xmax, ymax]. \\
			& Could you please extract and share the text within the defined coordinates [xmin, ymin, xmax, ymax]? \\
			& Please locate and provide the text that is encompassed by the given bounding box [xmin, ymin, xmax, ymax]. \\
			& Can you recognize and output the text enclosed within the specified coordinates [xmin, ymin, xmax, ymax]? \\
			& I'm looking for you to identify and deliver the text found within the provided bounding box [xmin, ymin, xmax, ymax]. \\
			& Could you extract and share the text within the defined coordinates [xmin, ymin, xmax, ymax], if available? \\
			& Please recognize and provide the text contained within the specified bounding box [xmin, ymin, xmax, ymax]. \\
			& Can you identify and furnish the text that falls within the provided coordinates [xmin, ymin, xmax, ymax]? \\
			\hline
			\multirow{10}{*}{spotting} & Could you locate the text in the image and furnish the coordinates [xmin, ymin, xmax, ymax] for each text block? \\
			& Please recognize all the text within the image and supply the coordinates [xmin, ymin, xmax, ymax] for each text element. \\
			& Can you identify and extract all the text from the image, and include the coordinates [xmin, ymin, xmax, ymax] for each text block? \\
			& I would like you to recognize the text within the image and provide the bounding box [xmin, ymin, xmax, ymax] for each piece of text. \\
			& Kindly identify and extract text from the image, and supply the coordinates [xmin, ymin, xmax, ymax] for each text portion. \\
			& Can you recognize all the text present in the image and provide the corresponding bounding boxes or coordinates [xmin, ymin, xmax, ymax]? \\
			& I'm looking for you to detect and list all text within the image, accompanied by their bounding box coordinates [xmin, ymin, xmax, ymax]. \\
			& Please analyze the image for text, and for each text segment, provide the bounding box coordinates [xmin, ymin, xmax, ymax]. \\
			& I'd appreciate it if you could identify and provide the coordinates [xmin, ymin, xmax, ymax] for all text found in the image. \\
			& Kindly pinpoint the text in the image and provide the coordinates [xmin, ymin, xmax, ymax] for each text block. \\
			\Xhline{2.5\arrayrulewidth}
		\end{tabular}
	}
	\caption{Instruction templates used for pre-training. These templates are divided into three parts, each corresponding to one of three tasks: detection, recognition, and spotting.}
	\label{tab:template}
\end{table*}

\noindent
\textbf{Fine-tuning.}
We employ the data generated by GPT-4 for fine-tuning our model.
Slightly different from the previous methods~\cite{liu2023visual,zhu2023minigpt,ye2023mplug,zhang2023llavar,feng2023unidoc,li2023blip}, our data focus on textual information present within images.
Moreover, when the response includes textual content, we instruct GPT-4 to create questions that encourage the model to provide reasoning related to the answer. 
This helps ensure the model to generate responses with textual bounding boxes.
Examples of such GPT-4 generated questions are shown in~\cref{tab:template_fine}.

\begin{table*}[!ht]
	\centering
	\resizebox{1.0\linewidth}{!}{
		\begin{tabular}{l}
			\Xhline{2.5\arrayrulewidth}
			Examples of GPT-4 generated questions in the conversation \\
			\hline
			What are the first few months that are featured on this calendar, along with their coordinates? \\
			What is the name of the boy mentioned in the image? Support your reasoning with the text and its bounding box. \\
			Can you share the event involving Nathan Lior, as well as its date and location? Could you provide the text and its bounding box which led to these answers? \\
			What kind of product is depicted in the image, and who made it? Please support your response with corresponding text and its bounding box. \\
			Can you infer the main topic of the magazine from the display texts? Can you also mention the bounding box for your reasoning? \\
			Locate any numbers present in the image along with their positions. Oh, and don't forget to provide the corresponding bounding boxes for your findings. \\
			Who has been recognized with the ``Best LegalTech Solution 2020'' award? Please specify the text and its bounding box to substantiate your response. \\
			Is there any noteworthy element in the image aside from these? Please support your answer with the corresponding text and its bounding box. \\
			What is the full phrase that contains the country name in the image? Please provide the text and its location. \\
			Who's the creative individual involved here? Also, could you indicate the corresponding text and its bounding box? \\
			Can you elaborate on the positioning of these texts? Please include the corresponding bounding boxes. \\
			Can I conclude this image pertains to a video game? Kindly justify your answer using the text and its associated bounding box. \\
			\Xhline{2.5\arrayrulewidth}
		\end{tabular}
	}
	\caption{Examples of GPT-4 generated questions for fine-tuning stage.}
	\label{tab:template_fine}
\end{table*}

\section*{B. Conversation Format of TGDoc}

\noindent
\textbf{System message.}
``A chat between a curious user and an artificial intelligence assistant. The assistant gives helpful, detailed, and polite answers to the user's questions.''

\noindent
\textbf{Conversation template of model.}
We provided one training template of the model as follows:

\textit{\textless System Message\textgreater USER: \textless image\textgreater \textless Image Embedding\textgreater \textless /image\textgreater What is the book's title based on the image? Please provide the supporting text and its bounding box. ASSISTANT: \textless Model's Output\textgreater}

\section*{C. The Processing of Data for Fine-tuning}
\begin{figure}[t]
	\centering
	\includegraphics[width=0.98\linewidth]{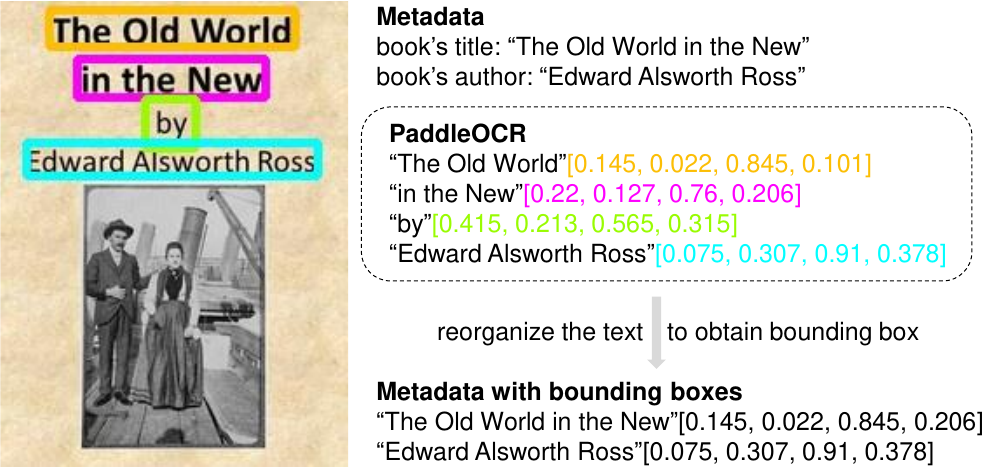}
	\caption{Acquisition of bounding boxes for the metadata of book cover. We employ an OCR tool to extract texts and position information from the image. Subsequently, the texts are reorganized to obtain the final bounding boxes corresponding to the metadata.}
	\label{fig:sup_fine}
\end{figure}

For the fine-tuning with book covers, we download both the images along with the associated metadata. 
However, the metadata lacks bounding box information.
To address this issue, we utilize PaddleOCR~\footnote{https://github.com/PaddlePaddle/PaddleOCR\label{paddle}} to extract text and corresponding bounding boxes from each cover.
Then, we reorganize the extracted text to constitute the metadata to obtain the final bounding boxes.
The entire process is shown in~\cref{fig:sup_fine}.

\section*{D. More Visual Results}
\begin{figure*}[t]
	\centering
	\includegraphics[width=0.98\linewidth]{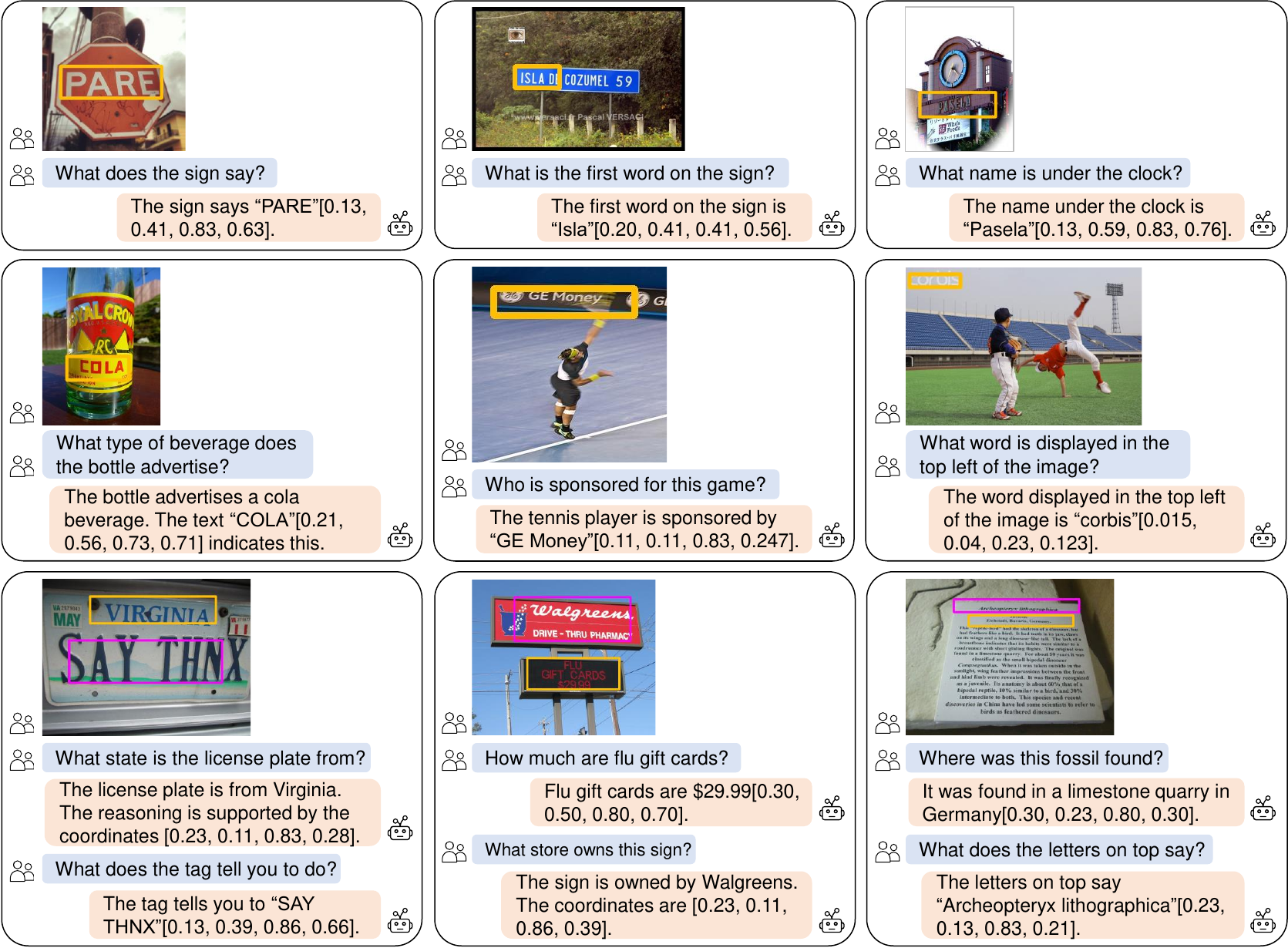}
	\caption{Visualization results of our method on STVQA~\cite{biten2019icdar} and TextVQA~\cite{singh2019towards} dataset.}
	\label{fig:more_results_1}
\end{figure*}

\begin{figure*}[t]
	\centering
	\includegraphics[width=0.98\linewidth]{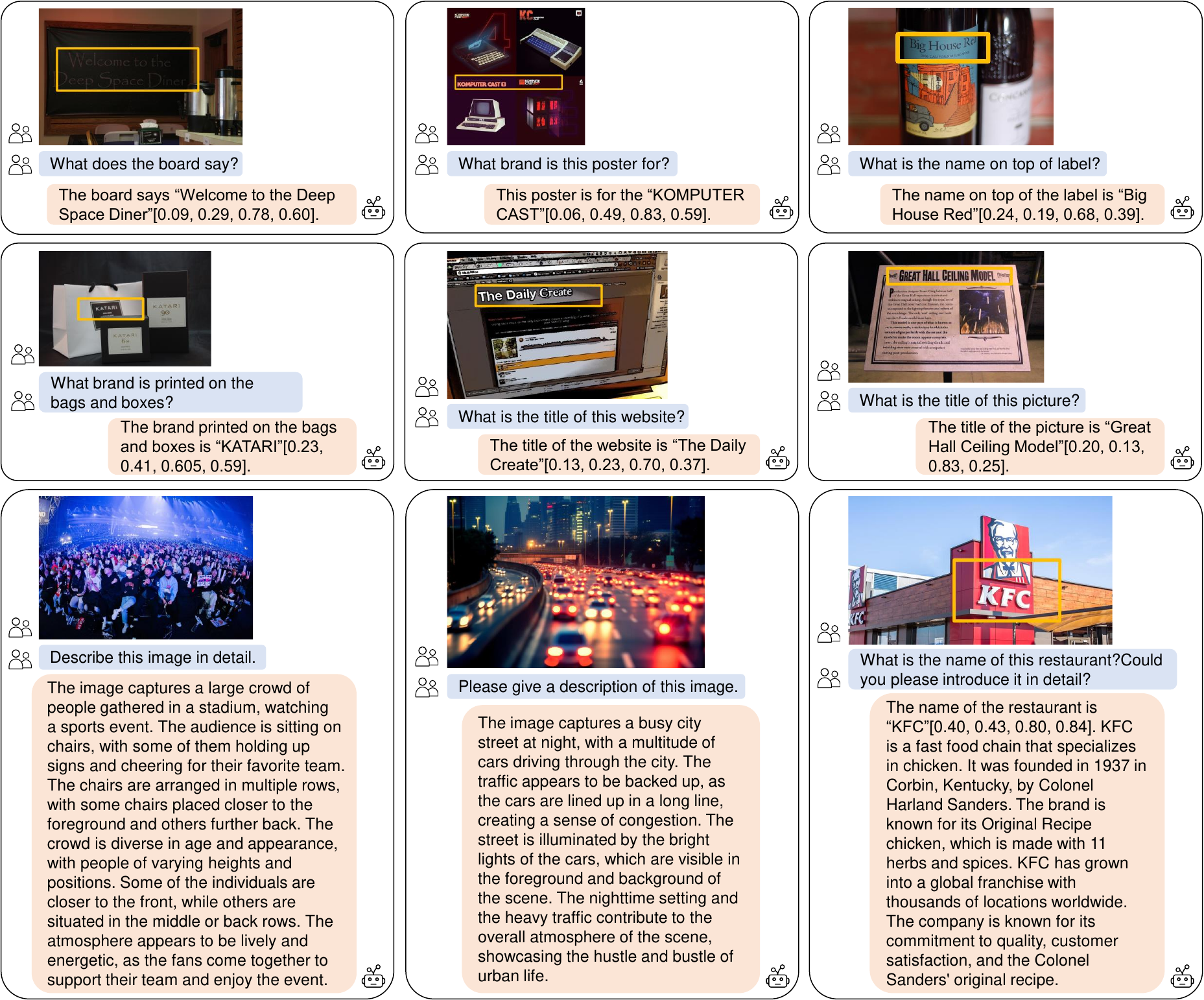}
	\caption{More visualization results of our method on STVQA~\cite{biten2019icdar} dataset, TextVQA~\cite{singh2019towards} dataset, and natural scenes.}
	\label{fig:more_results_2}
\end{figure*}

We present more visual results on STVQA~\cite{biten2019icdar} dataset, TextVQA~\cite{singh2019towards} dataset, and natural scenes in~\cref{fig:more_results_1} and ~\cref{fig:more_results_2}.
The results shows that the model is capable of determining the approximate locations of answers relative to the questions, thereby enabling precise extraction of text from images for model's response. 
Moreover, we have demonstrated TGDoc's proficiency in comprehending a variety of multimodal scenarios, further substantiating its robustness and adaptability in handling complex contexts. 
TGDoc effectively combines image content and textual data, improving its overall comprehension and accuracy across various scenarios.

{
    \small
    \bibliographystyle{ieeenat_fullname}
    \bibliography{main}
}